# CLASSIFICATION ANALYSIS OF AUTHORSHIP FICTION TEXTS IN THE SPACE OF SEMANTIC FIELDS

**Bohdan Pavlyshenko**

*Ivan Franko Lviv National University,*
Ukraine, pavlsh@yahoo.com

The use of naive Bayesian classifier (NB) and the classifier by the *k* nearest neighbors (kNN) in classification semantic analysis of authors' texts of English fiction has been analysed. The authors' works are considered in the vector space the basis of which is formed by the frequency characteristics of semantic fields of nouns and verbs. Highly precise classification of authors' texts in the vector space of semantic fields indicates about the presence of particular spheres of author's idiolect in this space which characterizes the individual author's style.

**Introduction**

Authors' texts represent a special author's idiolect that characterizes a set of author's means of expression, particularly a distinctive semantic range of author's lexicon. The use of quantitative intellectual analysis enables to research the features of author's style and to analyse the authorship of unknown texts. Widely used in the text mining algorithms is a vector model where text documents are represented as vectors in some phase space [1]. This model is often used in classification analysis of texts [1,2,3]. One of the problems is a large dimension of the vector space. Timely is to find an effective basis of the vector space of text documents for the classification and cluster analyses. Classification analysis can be used not only for searching semantically similar documents, but also for analyzing the author's style. In the problems of analyzing text content analyzing a theory of lexical semantics is relevant, including the study of semantic fields. An example of lexicographic computer system, where a semantic network of links among lexemes is represented, is WordNet (http://wordnet.princeton.edu), developed in Princeton University [4]. This system is based on expert lexicographic analysis of semantic structural relationships that reflect the denotative and connotative characteristics of the vocabulary lexeme structure. The semantic structure of vocabulary lexemes can be used in the relevant classification and clustering algorithms for text objects in terms of reducing the dimensionality problem of analysis and identification of new semantic features. Databases WordNet are created by expert lexicographers. Nouns, verbs, adjectives and adverbs are organized in synsets – sets of synonyms. Nouns and verbs are grouped according to semantic fields. In [5] the concept of semantic domain is introduced, which describes certain semantic areas of various issues discussed, such as economics, politics, physics, programming, etc.

In our work we consider the use of vector space of semantic fields in the classification analysis of authors' texts in fiction. We also analyze the precision and recall of naive Bayesian classifier and the k nearest neighbors classifier for a vector model of text documents in the space of semantic fields.

**The Classifiers of Text Documents in the Space of Semantic Fields**
A set of text documents we describe as:

$$D = \{d_j \mid j = 0,1,2..., N_d\}. \tag{1}$$

Let us introduce a set of semantic fields

$$S = \{s_k \mid k = 1,2..., N_s\}. \tag{2}$$

Semantic field is a set of lexemes which are united by some common concept [4,5]. An example of semantic fields can be the field of motion, the field of communication, the field of perception, etc. Let there is some vocabulary of lexemes that occur in text areas $W = \{w_i \mid i = 1,2..., N_w\}$
We define the lexeme structure of semantic fields $S_k$ as

$$W_k^s = \left\{ w_i \mid w_i \xrightarrow{U_{ws}} s_k, i = 1,2..., N_w \right\}. \tag{3}$$

We introduce the frequency of a semantic field as

$$p_{kj}^{sd} = \sum_{i=1}^{N_w} p_{ij}^{wd} f_s(w_i, s_k), \quad f_s(w_i, s_k) = \begin{cases} 1, w_i \in W_k^s \\ 0, w_i \notin W_k^s \end{cases}, \tag{4}$$

where we write $p_{ij}^{wd}$ as the lexeme text frequency $w_i$ in the document $d_j$, which is defined by the ratio of given number of lexemes $w_i$ to the total number of lexemes in the document $d_j$. The combination of the values $p_{kj}^{sd}$ form a matrix of a feature-document type where the features are the frequencies of semantic fields in the documents:

$$M_{sd} = \left( p_{kj}^{sd} \right)_{k=1, j=1}^{N_s, N_d}. \tag{5}$$

The vector

$$V_j^s = \left( p_{1j}^{sd}, p_{2j}^{sd}, ..., p_{N_s j}^{sd} \right) \tag{6}$$

describes the document $d_j$ in $N_s$-dimensional space of text documents with the basis formed by semantic fields.

Let there are some categories of text documents. In our analysis such categories are formed by texts grouped by authors. The set of these categories we denote as

$$Categories = \{ Ctg_m \mid m = 1,2,..., N_{ctg} \}, \tag{7}$$

where $N_{ctg} = |Categories|$ defines the size of categories set. According to given categories text documents of the D (1) set are distributed. The goal is in finding the target function, which is described as

$$F_{d \to ctg} : Categories \times D \to \{0,1\} \tag{8}$$

Let us consider the naive Bayesian classifier of text documents. In the existing methods of text classification based on the naive Bayesian classifier, the frequency of corresponding keywords is used for documents presentation [2,3]. The approach which is based on the documents presentation by frequency characteristics of semantic fields, is promising because of lower dimension of the semantic vector space. Let find the a posteriori probability that due to some set of semantic fields frequencies the document $d_j$ belongs to the category $ctg_m$. Using the Bayes theorem we define that

$$P(ctg_m \mid p_{1j}^{sd}, p_{2j}^{sd}, ..., p_{N_s j}^{sd}) = \frac{P(ctg_m) P(p_{1j}^{sd}, p_{2j}^{sd}, ..., p_{N_s j}^{sd} \mid ctg_m)}{P(p_{1j}^{sd}, p_{2j}^{sd}, ..., p_{N_s j}^{sd})} \tag{9}$$

In the implementation the naive Bayesian classificator researches make a significant assumption about conditional independence of attributes of objects [2]. In this case the conditional probability $P(p_{1j}^{sd}, p_{2j}^{sd},..., p_{N_s j}^{sd} | ctg_m)$ is approximated by the product of conditional probabilities $P(p_{ij}^{sd} | ctg_m)$. Continuous distributions $P(p_{ij}^{sd} | ctg_m)$ are often approximated by normal Gaussian distribution. As the parameters of this distribution, one may consider the mean and variance of semantic fields frequencies. A rule of decision making about the inclusion of analyzed document to a particular category is an addition to the calculation of naive Bayesian classifier. In the simplest case of such a rule may decide about document belonging to a given category, if calculated a posteriori probability for such a category at given semantic field frequencies is the largest, that is

$$Category(d_j) = ctg_m : P(ctg_m | p_{1j}^{sd}, p_{2j}^{sd},..., p_{N_s j}^{sd}) = $$
$$= \max\{P(ctg_k | p_{1j}^{sd}, p_{2j}^{sd},..., p_{N_s j}^{sd}) | k = 1,2,...., N_{ctg}\} \quad (10)$$

The probabilities $P(p_{ij}^{sd} | ctg_m)$ are being formed on some educational categorized array of text documents.

Now we consider the classification by the k nearest neighbors, which is called the kNN classification [2,3]. This method is referred to as vector classifiers. The basis of vector methods of classification is the hypothesis of compactness. According to this hypothesis, the documents belonging to one and the same class create a compact area, and areas that belong to different classes do not intersect. As a measure of closeness among documents we choose Euclidean distance. In kNN classification the boundaries of categories are defined locally. Some document refer to a category, which is dominant for $k$ its neighbors. If $k = 1$, a document attributes the class of its nearest neighbor. Under the hypothesis of compactness a test document $d$ has the category that most documents in the training sample have in some local spatial neighborhood of a document $d$.

Let us consider the estimation of precision of documents classification [2,3]. Classifier's decision-making about the document $d_j$ belonging to the category $ctg_i$ we mark as $Class(d_i) = Ctg_j$. A set of documents identified by classifier as appropriate to the category $ctg_i$ and they really belong to this category due to expert review, should look

$$Set_1^{tclass} = \{d_i | Class(d_i) = Ctg_j \land d_i \in Ctg_j\}. \quad (11)$$

A set of documents defined by classifier as belonging to the category $ctg_i$ looks like

$$Set_2^{tclass} = \{d_i | Class(d_i) = Ctg_j\}. \quad (12)$$

A set of documents that belong to the category $ctg_i$ looks like

$$Set_3^{tclass} = \{d_i | d_i \in Ctg_j\}. \quad (13)$$

Precision and recall are used to characterize classifiers. The precision of the classifier is defined as the ratio of the number of elements of the set $Set_1^{tclass}$ to the elements of the set $Set_2^{tclass}$

$$\Pr_j^{tclass} = \frac{Set_1^{tclass}}{Set_2^{tclass}} = \frac{|\{d_i | Class(d_i) = Ctg_j \land d_i \in Ctg_j\}|}{|\{d_i | Class(d_i) = Ctg_j\}|} \quad (14)$$

Recall is defined as the ratio of the number of elements of the set $Set_1^{tclass}$ to the elements of the set $Set_3^{tclass}$

$$Rc_j^{tclass} = \frac{Set_1^{tclass}}{Set_3^{tclass}} = \frac{|\{d_i \mid Class(d_i) = Ctg_j \land d_i \in Ctg_j\}|}{|\{d_i \mid d_i \in Ctg_j\}|} \quad (15)$$

In the characteristics $\Pr_j^{tclass}$, $Rc_j^{tclass}$ index j defines the category and index tclass determines the type of a classifier. In our studies

$$tclass = \{NB, nKNN\} \quad (16)$$

Each category of the documents is characterized by its values $\Pr_j^{tclass}$, $Rc_j^{tclass}$. For the general characteristics of the classifier we find the macro-averaging of indicators $\Pr_j^{tclass}$, $Rc_j^{tclass}$ by all categories

$$\Pr_{mean}^{tclass} = \frac{1}{N_{ctg}} \sum_{i=1}^{N_{ctg}} \Pr_i^{tclass}, \quad (17)$$

$$Rc_{mean}^{tclass} = \frac{1}{N_{ctg}} \sum_{i=1}^{N_{ctg}} Rc_i^{tclass}. \quad (18)$$

**Experimental Part**

For the experimental study of text documents classification in the space of semantic fields we chose a text base containing 503 literary works of 17 authors. For the semantic space generation we chose the lexemes grouped by the semantic fields of nouns and verbs in the semantic network WordNet [4]. The semantic fields in the WordNet network (http://wordnet.princeton.edu) are represented by lexicographic files. In our studies we have used the semantic fields of nouns and verbs. The semantic fields of nouns consist of 26 lexicographic files out of which we have selected 54464 tokens. The semantic fields of verbs contain 15 lexicographic files into which we have selected 9097 tokens. The derivative forms of lexemes were also included into the semantic fields. Lexicographic files WordNet for nouns and verbs have the names that define the semantic core of these fields: noun.Tops, noun.act, oun.animal, noun.artifact, noun.attribute, noun.body, noun.cognition, noun.communication, noun . event, noun.feeling, noun.food, noun.group, noun.location, noun.motive, noun.object, noun.person, noun.phenomenon, noun.plant, noun.possession, noun.process, noun.quantity , noun.relation, noun.shape, noun.state, noun.substance, noun.time, verb.body, verb.change, verb.cognition, verb.communication, verb.competition, verb.consumption, verb.contact, verb.creation, verb.emotion, verb.motion, verb.perception, verb.possession, verb.social, verb.stative, verb.weather.

With the developed software the initial processing of text array is made, the auxiliary symbols and text elements without semantic information were removed. For each document and the entire sample, the frequency dictionaries were calculated, on the basis of which we have calculated the matrix $M_{sd}$ (5) of the type document- frequency_of_semantic_field. Based on this matrix, we investigated two types of classifiers – naive Bayesian classifier (NB) and the classifier by the *k* nearest neighbors (kNN). Let us consider the main results obtained. The training set included 350 documents and the test one contained 153 documents selected randomly. The precision and recall of Bayesian classifier for text documents of different authors are shown in Fig. 1.

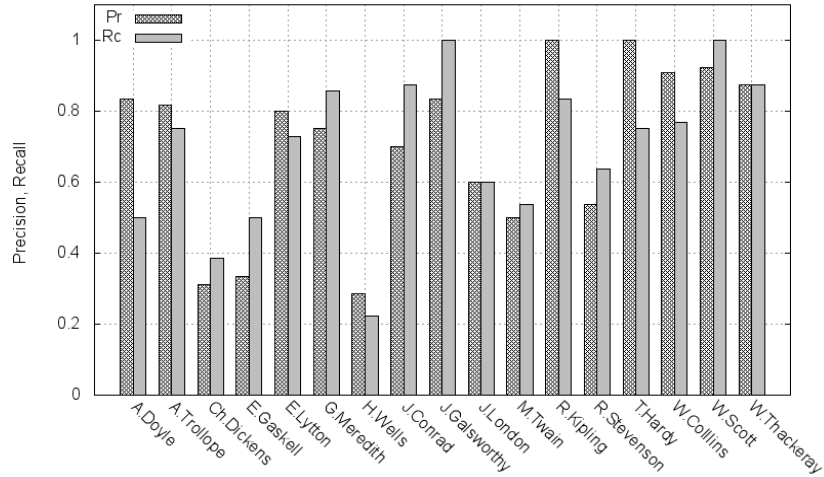

Fig. *1. Precision and recall of Bayesian classifier.*

Precision and recall of the classifier by the nearest k neighbors when k = 5 for text documents by different authors are shown in Fig. 2.

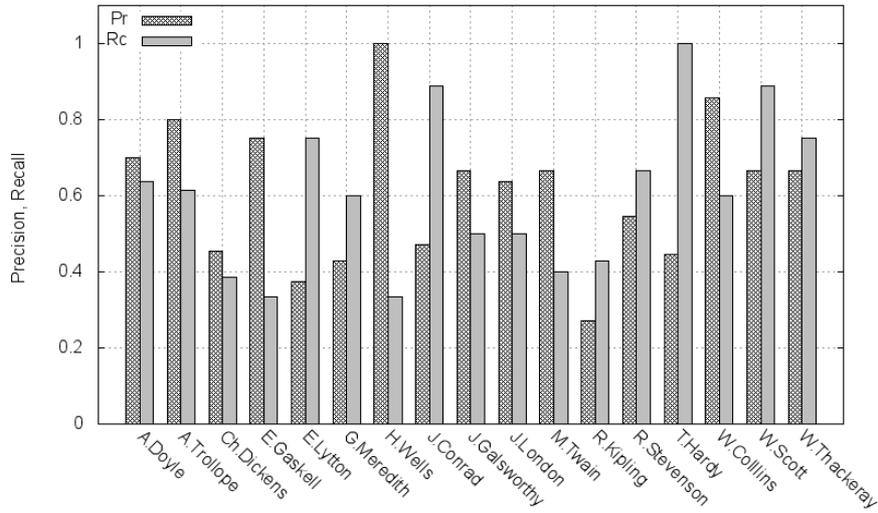

*Fig.2 Precision and recall of the classifier by the nearest k neighbors (k = 5).*

For Bayesian classification in all fields we obtained the following macro parameters: $Pr_{mean}^{tclass} = 0.7066$, $Rc_{mean}^{tclass} = 0.6952$. In case of Bayesian classification by nouns only $Pr_{mean}^{tclass} = 0.6135$, $Rc_{mean}^{tclass} = 0.6295$. In case of Bayesian classification by verbs only $Pr_{mean}^{tclass} = 0.5834$, $Rc_{mean}^{tclass} = 0.5539$. In case of Bayesian classification when training and test samples are identical and contain 503 documents, we obtained $Pr_{mean}^{tclass} = 0.8421$, $Rc_{mean}^{tclass} = 0.8540$. In case of kNN classification when k = 5 we obtained $Pr_{mean}^{tclass} = 0.6119$, $Rc_{mean}^{tclass} = 0.6045$. In case of kNN classification when k = 1 we obtained $Pr_{mean}^{tclass} = 0.5748$, $Rc_{mean}^{tclass} = 0.6071$. The results given in Fig. 1, 2 show varying effectiveness of NB and kNN classifiers for texts of certain authors. Thus, for the authors G.Meredith, T.Hardy, R.Kipling classifier NB shows high precision, and kNN shows low precision. For the authors E.Gaskell, H.Wells classifier kNN shows high precision, and NB vice versa - low precision.

**Conclusion**

We have analyzed the use of naive Bayesian classifier (NB) and the classifier by the *k* nearest neighbors (kNN) in classification semantic analysis of authors' texts of English fiction. The authors' works are considered in the vector space, the basis of which is formed by the frequency characteristics of semantic fields of nouns and verbs. The dimension of such a basis is significantly lower than the one in case of commonly used classification of texts by keywords. The results obtained demonstrate the effectiveness of NB and kNN classification in the space of semantic fields. Categorical distributions of precision and recall can vary widely for different classifiers with one and the same training and test samples. Highly precise classification of authors' texts in the vector space of semantic fields indicates about the presence of particular spheres of author's idiolect in this space which characterizes the individual author's style. The effectiveness of classification analysis of authors' texts in the space of semantic fields increases if to combine different classification methods particularly Bayesian classification and the classification by the *k* nearest neighbors.